\pdfoutput=1

\documentclass[11pt]{article}

\usepackage[]{acl}

\usepackage{times}
\usepackage{latexsym}
\usepackage{graphicx}
\usepackage{arydshln}
\usepackage{amsmath}
\usepackage{booktabs}
\usepackage{multirow}
\usepackage{float}
\usepackage[T1]{fontenc}
\newcommand{\st}{Statement-Tuning}

\usepackage[utf8]{inputenc}

\usepackage{microtype}

\newcommand{\rparagraph}[1]{\vspace{1.4mm}\noindent\textbf{#1.}}

\title{Enabling Natural Zero-Shot Prompting on Encoder Models via \st}

\author{Ahmed Elshabrawy\textsuperscript{\textnormal{1}} ,
Yongxin Huang\textsuperscript{\textnormal{2}} ,
Iryna Gurevych\textsuperscript{\textnormal{1,2}}, 
Alham Fikri Aji\textsuperscript{\textnormal{1}} \\
\textsuperscript{1}Department Natural Language Processing, MBZUAI\\
\textsuperscript{2}Ubiquitous Knowledge Processing Lab, Technical University of Darmstadt\\
\small{\textsuperscript{1}\texttt{\{ahmed.elshabrawy,iryna.gurevych,alham.fikri\}@mbzuai.ac.ae}} \\
\small{\textsuperscript{2}\texttt{ www.ukp.tu-darmstadt.de}}}

\begin{document}
\maketitle
\begin{abstract}
While Large Language Models (LLMs) exhibit remarkable capabilities in zero-shot and few-shot scenarios, they often require computationally prohibitive sizes. Conversely, smaller Masked Language Models (MLMs) like BERT and RoBERTa achieve state-of-the-art results through fine-tuning but struggle with extending to few-shot and zero-shot settings due to their architectural constraints. Hence, we propose \st, a technique that models discriminative tasks as a set of finite statements and trains an encoder model to discriminate between the potential statements to determine the label. We do \st~on multiple tasks to enable cross-task generalization. Experimental results demonstrate that \st~achieves competitive performance compared to state-of-the-art LLMs with \textit{significantly} fewer parameters. Moreover, the study investigates the impact of several design choices on few-shot and zero-shot generalization, revealing that \st~can achieve strong performance with modest training data and benefits from task and statement diversity for unseen task generalizability. We release all the code used to generate statement data, train and evaluate our Statement-Tuned models.\footnote{Work In Progress}

\end{abstract}

\section{Introduction}
Large Language Models (LLMs) have shown great capabilities in zero-shot and few-shot settings \cite{radford-etal-2019-multitask, brown-etal-2020-fewshot, artetxe-etal-2022-efficient}. 
However, such capabilities are more difficult to observe in encoder-only models like BERT \cite{devlin-etal-2019-bert} and RoBERTa \cite{liu2019roberta} due to their architectural design. 
These models are typically pretrained in an unsupervised manner on a large corpus with a Masked Language Modeling \cite{devlin-etal-2019-bert} or Discriminative \cite{clark2020electra} objective and fine-tuned by adding task-specific layers to enable their usage on a particular task such as binary/multi-label classification, token/sequence classification, multiple choice, etc. These task-specific layers, thus, can not be extended effectively to new tasks in a few-shot or zero-shot manner.

In this work, we explore the feasibility of utilizing encoder models that are usually specialized for a certain task to take on various, unseen Natural Language Understanding (NLU) tasks, akin to zero-shot prompting in decoder models. One benefit of using encoder models is that they are generally more compact. Yet, encoder models have achieved state-of-the-art results on many NLU tasks through task-specific fine-tuning. So it would be interesting if zero-shot/few-shot prompting could be adapted for encoder models to leverage their powerful NLU capabilities at more computationally feasible sizes.

To achieve this, some techniques try to reformulate various downstream tasks with a unified format resembling the pre-training objective, enabling few-shot transfer for encoder models \cite{schick-schutze-2021-exploiting, schick-schutze-2021-just, xia-etal-2022-prompting}. Without few-shot examples, the zero-shot generalization of these models relies mainly on the language modelling ability learned in the pre-training phase, not benefiting from further multitask training on diverse reformulated tasks. 

In this work, we take inspiration from multitask instruction tuning methods for decoder models \cite{DBLP:conf/iclr/WeiBZGYLDDL22, sanh2022multitask} 
and unified format fine-tuning methods for encoder models \cite{xu-etal-2023-universal}
to propose \st, a novel intuitive approach for encoder-only models to generalize to zero-shot and few-shot unseen tasks through universal multitask fine-tuning with data formatted as statements. Our approach thus has the generalization ability similar to decoder models with a fraction of the parameters and training data.
\begin{figure*}
    \centering
    \includegraphics[width=0.8\linewidth]{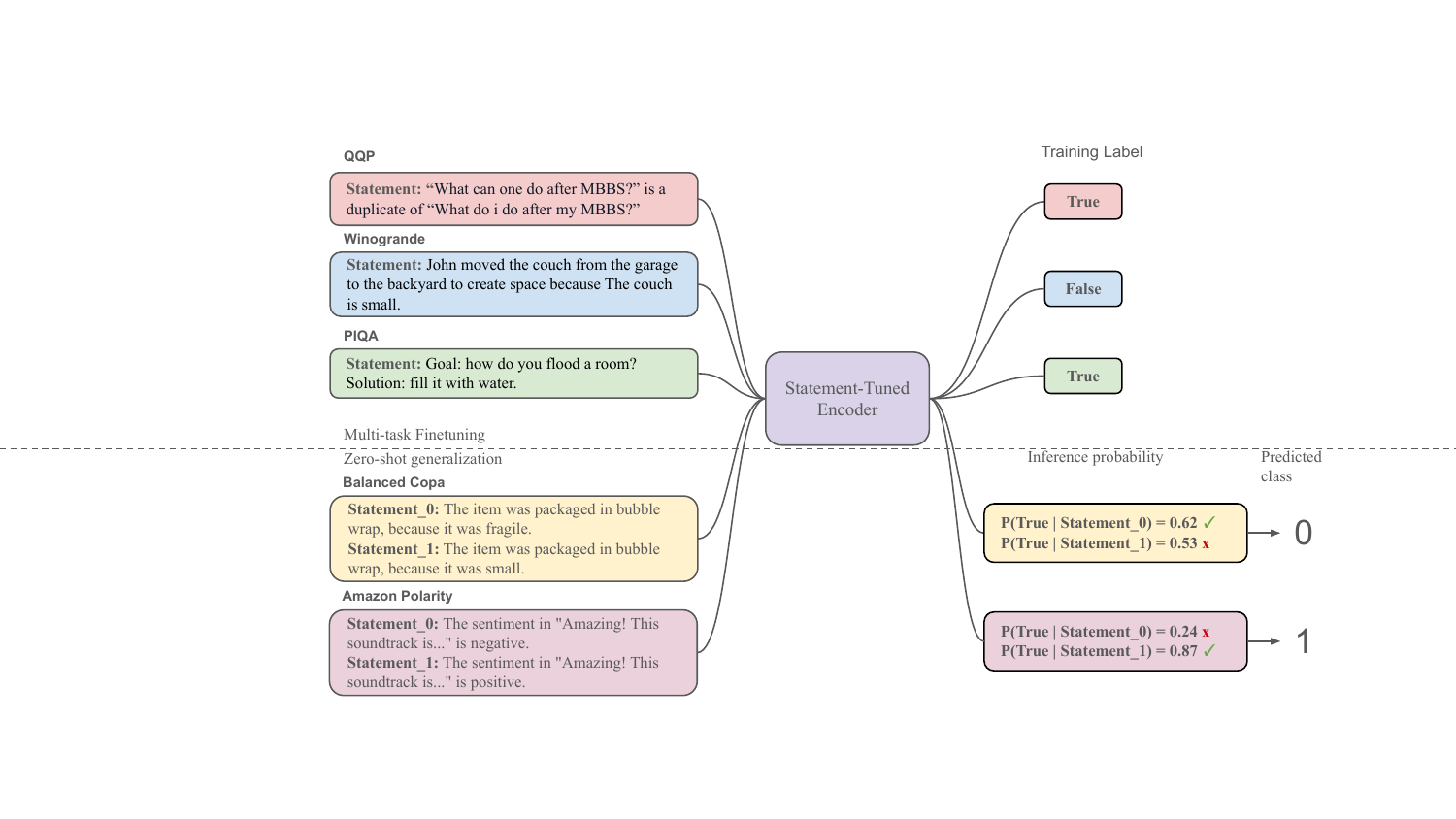}
    \caption{Overview of \st. We train an encoder to discriminate the truth value of statements from multiple tasks, then we apply it in the zero-shot setting by creating a statement for each possible target label and choosing the most likely one according to the encoder discriminator.}
    \label{fig:statement_tuning}
\end{figure*}

As seen in Figure~\ref{fig:statement_tuning}, we verbalize a diverse set of NLU tasks into natural language statements, and then fine-tune an encoder-only MLM, RoBERTa, on a universal binary sequence classification task, which we call \st, to assign a truth value (True or False) to any given statement.
By fine-tuning encoder models across diverse tasks and statements, we show zero-shot generalization capabilities to unseen tasks by similarly transforming them into statements. Moreover, we show few-shot capabilities by continually fine-tuning this model with a small amount of downstream data, also formatted into statements. \st~is capable of matching or even outperforming (32-shot and) zero-shot performance of many state-of-the-art LLMs with a fraction of the parameters.

Our ablation study shows that depending on the task, we can achieve substantial few-shot and zero-shot generalizability with as few as 1,000 statements per training dataset or approximately 16,000 training statement examples in total, which correspond to even fewer original task examples since one example can be turned into multiple statements through different templates. Furthermore, we find that 
the statement and task diversity tend to have a beneficial effect on the performance and generalizability of \st. 

In summary, our primary contributions are:

\begin{enumerate}
\itemsep0em
    \item To the best of our knowledge, we are the first to propose a combination of elaborate statement formulation and Masked Language Models as a simple and effective data/resource-efficient alternative for LLMs for zero-shot NLU task generalization.
    \item We expose that the emergent ability \citep{wei2022emergent} to solve unseen tasks after training on a large volume of instruction-tuning data, which is previously thought to be exclusive to decoder-based LLMs of large-enough sizes, can also be observed in much smaller MLMs when we do multitask \st~with a modest amount of training data.
    \item We explore a large number of design choices to study how \st~benefits from the number of statement examples and statement template and task diversity in multitask \st, and demonstrate the data/resource-efficiency of~\st.

\end{enumerate}



\section{Related Work}
\paragraph{Zero-Shot and Few-Shot Approaches Utilizing Label Semantics}
Various approaches have been proposed to reformulate zero-shot classification to leverage label semantics instead of indices, enabling the more generalized use of encoder models. In the reformulated task, textual labels are combined with the original input text, and the model should predict whether the label matches the text. TARS \cite{halder-etal-2020-task} utilizes the simple concatenation of input text and label text, with no explicit connection through natural language. Similarly, \citet{xu-etal-2023-universal} reformulate discriminative tasks with minimal prompts, which are mostly simple concatenations of elements in the raw input. They find it effective for zero-shot generalization for ELECTRA-style encoder models but don’t test other encoder-only models with normal MLM training. 
\citet{yin-etal-2019-benchmarking} propose an entailment-based approach where the input is a pair of texts: the original input text is the premise and label is converted into a hypothesis. However, the approach is limited by the fact that not every discriminative task can be formulated as an entailment task (in the form of a premise and a hypothesis). The entailment formulation also limits the training tasks to entailment datasets (MNLI, RTE, and FEVER), which often causes the model to over-rely on spurious lexical patterns, hindering generalization \citep{ma-etal-2021-issues}. Our approach \st~serves as a more universal formulation of any discriminative task in the form of statements.

Few-shot approaches utilizing cloze-style templates have also been proposed. PET \cite{schick-schutze-2021-exploiting} requires an ensemble of encoder models and iterative training, relies on additional unlabeled data; Improved PET variants \cite{schick-schutze-2021-just, tam-etal-2021-improving} use complex losses to compute the probability of each token in multi-token labels. Our method keeps the normal sequence classification training paradigm, while in the meantime effectively leverages the label semantics, enabling straightforward zero-shot generalization on a single encoder model.

\paragraph{Zero-Shot Prompting and Multitask Tuning}
LLMs excel at unseen-task/zero-shot generalization \cite{brown-etal-2020-fewshot}. Building on this, recent work explores multitask training with diverse prompts for improved zero-shot performance \cite{sanh2022multitask, DBLP:conf/iclr/WeiBZGYLDDL22, chung2022scaling}. These methods fine-tune large models on constructed datasets with various task prompts, achieving strong zero-shot results on unseen tasks. However, effective instruction-tuning often requires billions of parameters \cite{zhang2024instruction}, limiting their application to smaller models. \citet{ye-etal-2022-zerogen} aim to distill this zero-shot ability in a smaller model like an LSTM through synthetic data creation using an LLM, but they create task-specific models rather than a single smaller model that is capable of generalizing. Our work demonstrates similar or superior generalization than LLMs using a single smaller MLM with less training data.
\begin{figure}[!htp]
    \centering
    \includegraphics[width=0.9\linewidth]{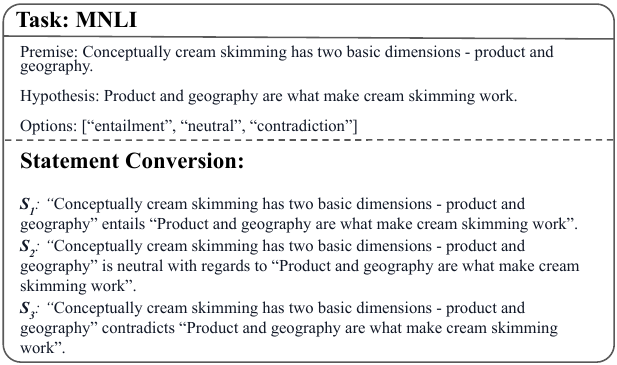}
    \caption{Example conversion of the MNLI task to natural language statements.}
    \label{fig:conversion}
\end{figure}

\section{Method: \st}
In this section, we outline the steps involved in \st. First, tasks are verbalized into natural language statements. Then they are used to train the statement discriminator and derive the target label.

\paragraph{Task Verbalization}
Any discriminative task with a finite set of targets can be verbalized into a finite set of natural language statements. Figure~\ref{fig:conversion} shows the example of converting the MNLI task into statements. Similar to prompting, each task has its own statement templates, based on each possible label. The truth label for training purposes on each statement depends on whether the statement contains the correct target label or not. 


\paragraph{Statement Fine-Tuning}
\label{sec:st&a}

To create the training data for multitask statement fine-tuning, we exhaustively generate statements across 16 diverse NLP datasets (categorized into 9 tasks, see Appendix \ref{sec:appendixe}) using many varied statement templates (see Appendix \ref{sec:appendixa}) per dataset: QQP \cite{sharma2019natural}, Winogrande \cite{ai2:winogrande}, PiQA \cite{Bisk2020}, MNLI \cite{williams-etal-2018-broad}, SNLI \cite{bowman-etal-2015-snli}, Mintaka \cite{sen-etal-2022-mintaka}, Yelp Polarity \cite{zhangCharacterlevelConvolutionalNetworks2015}, WikiLingua \cite{ladhak-etal-2020-wikilingua}, SQuAD \cite{2016arXiv160605250R}, TweetEval's Offensive task \cite{zampieri2019semeval}, Massive \cite{fitzgerald2022massive, bastianelli-etal-2020-slurp}, Definite Pronoun Resolution \cite{rahman2012resolving}, QASC \cite{allenai:qasc}, SciQ \cite{SciQ}, RACE \cite{lai-etal-2017-race}, and SAMSum \cite{gliwa-etal-2019-samsum}.
We fine-tune RoBERTa \cite{liu2019roberta} with a binary sequence classification head to predict the truth value of the statements. By fine-tuning the model across diverse tasks, templates, and domains, the model should be able to generalize across unseen templates and tasks, as long as it can be phrased as a true/false statement.



\paragraph{Zero-Shot and Few-Shot Inference}

To perform inference on statement-finetuned RoBERTa, we also need to transform the input into statements. We randomly choose a statement template for each dataset at inference time. In our experiments, we show that our statement-tuning is robust to different templates. We exhaustively generate a statement for each possible label, as shown in Figure~\ref{fig:statement_tuning}.
Then, for each statement corresponding to each label, we predict the probability of such a statement being true. The final label is the statement with the highest true probability. Zero-shot inference is done by directly performing the aforementioned inference regime on the statement-finetuned RoBERTa, while K-shot inference is done after continual fine-tuning on K examples of task-specific statements.


\begin{table*}[!ht]
    \centering
    \resizebox{\textwidth}{!}{
    \begin{tabular}[width=\linewidth]{@{ }l@{ }ccccccccc}
    \toprule
         &  \textbf{\#Parameters} &   \multicolumn{1}{c}{\textbf{BCOPA}}   &   \multicolumn{1}{c}{\textbf{MRPC}}   &   \multicolumn{1}{c}{\textbf{FigQA}}   &   \multicolumn{1}{c}{\textbf{Amazon Polarity}}   &   \multicolumn{1}{c}{\textbf{StoryCloze}}   &   \multicolumn{1}{c}{\textbf{YA Topic}}   &   \multicolumn{1}{c}{\textbf{Emotion}} &  \multicolumn{1}{c}{\textbf{Avg}} \\
         \midrule
         Meta-Llama-3-70B-Instruct &70B &89.0 &\textcolor{lightgray}{\textbf{71.3}} &\textcolor{lightgray}{\textbf{42.0}} &94.7 &82.7 &61.9 &51.8 & 70.5 \\
        Llama-2-13b-chat-hf &13B &89.6 &60.8 &\textcolor{lightgray}{\textbf{40.9}} &93.7 &82.4 &53.2 &51.6 & 67.5 \\
         Llama-2-7b-chat &7B &86.6 &\textcolor{lightgray}{\textbf{54.4}} &\textcolor{lightgray}{\textbf{40.1}} &\textcolor{lightgray}{\textbf{90.5}} &\textcolor{lightgray}{\textbf{78.5}} &47.8 &50.0 & 64.0 \\
    Mistral-7B-Instruct-v0.2 &7B &89.4 &73.0 &\textcolor{lightgray}{\textbf{41.4}} &\textcolor{lightgray}{\textbf{88.9}} &\textcolor{lightgray}{\textbf{82.3}} &57.7  &55.3 & 69.7 \\
    Qwen1.5-7B-Chat &7B &87.0 &75.5 &\textcolor{lightgray}{\textbf{42.1}} & 95.3 & 79.7 &59.1 &57.8 & 70.9 \\
    Pythia-6.9B &6.9B &82.2 &\textcolor{lightgray}{\textbf{62.0 }} &\textcolor{lightgray}{\textbf{41.7}} &\textcolor{lightgray}{\textbf{83.3}} &\textcolor{lightgray}{\textbf{71.2}} &\textcolor{lightgray}{\textbf{32.2}} &\textcolor{lightgray}{\textbf{25.1}} & 56.8 \\
    Pythia-2.8B &2.8B &79.6 &\textcolor{lightgray}{\textbf{68.4}} &\textcolor{lightgray}{\textbf{41.2}} &\textcolor{lightgray}{\textbf{77.7}} &\textcolor{lightgray}{\textbf{69.7}} &\textcolor{lightgray}{\textbf{12.1}} &\textcolor{lightgray}{\textbf{35.4}} & 54.9 \\
    Phi-2 &2.7B &87.2 &\textcolor{lightgray}{\textbf{67.9}} &\textcolor{lightgray}{\textbf{41.8}} &\textcolor{lightgray}{\textbf{86.6}} &\textcolor{lightgray}{\textbf{77.7}} &\textcolor{lightgray}{\textbf{38.7}} &53.1 & 64.7 \\
    FlanT5-Large &770M &\textcolor{lightgray}{\textbf{67.6}} &81.1 &\textcolor{lightgray}{\textbf{40.1}} &96.0 &\textcolor{lightgray}{\textbf{63.0}} &51.0 &59.9 & 65.5 \\
    Qwen1.5-0.5B-Chat &500M &\textcolor{lightgray}{\textbf{69.2}} &\textcolor{lightgray}{\textbf{32.6}}  &\textcolor{lightgray}{\textbf{38.7}} &\textcolor{lightgray}{\textbf{69.7}} &\textcolor{lightgray}{\textbf{68.9}} &\textcolor{lightgray}{\textbf{21.9}} &\textcolor{lightgray}{\textbf{6.6}} & 43.9 \\
    BART-large-mnli &406M  &\textcolor{lightgray}{\textbf{50.4}} &\textcolor{lightgray}{\textbf{35.8}}  &\textcolor{lightgray}{\textbf{46.9 }} &\textcolor{lightgray}{\textbf{49.4}} &\textcolor{lightgray}{\textbf{47.3}}  &\textcolor{lightgray}{\textbf{6.5}} &\textcolor{lightgray}{\textbf{11.7}} & 35.4 \\
    FlanT5-Small &60M  &\textcolor{lightgray}{\textbf{52.8 }} &\textcolor{lightgray}{\textbf{31.9}}  &\textcolor{lightgray}{\textbf{42.0}}  &\textcolor{lightgray}{\textbf{88.8}}  &\textcolor{lightgray}{\textbf{51.5}}  &\textcolor{lightgray}{\textbf{24.5}}  &\textcolor{lightgray}{\textbf{21.7}}& 44.7 \\
    \hdashline
    Our Approach: \\
    \textbf{RoBERTa-base (Best)} &\textbf{125M} &75.3$_{(0.5)}$ &72.3$_{(1.5)}$ &61.4$_{(0.6)}$ &92.9$_{(1.3)}$ &79.1$_{(1.1)}$ &40.2$_{(3.8)}$ &48.5$_{(5.1)}$ & 67.1 \\
    \textbf{RoBERTa-base (4k)} &\textbf{125M} &72.4$_{(0.5)}$ &69.6$_{(1.1)}$ &60.7$_{(0.9)}$ &92.3$_{(0.8)}$ &78.5$_{(2.7)}$ & 37.9$_{(2.7)}$ &46.6$_{(4.3)}$ & 65.4 \\
    \textbf{RoBERTa-large (Best)} &\textbf{355M} &85.1$_{(0.7)}$ &71.8$_{(0.8)}$ &74.2$_{(1.4)}$ &95.4$_{(0.4)}$ & 92.1$_{(0.7)}$ &49.9$_{(2.1)}$ &50.7$_{(1.4)}$ & 75.3 \\
    \textbf{RoBERTa-large (10k)} &\textbf{355M} &85.1$_{(0.7)}$ &71.5$_{(0.8)}$ &73.0$_{(2.4)}$ &95.4$_{(0.4)}$ &91.1$_{(0.8)}$ &48.4$_{(0.7)}$ &49.1$_{(3.2)}$ & 73.4 \\
    \midrule
    Full/3000-shot:\\
    RoBERTa-base (FT) &125M &\multicolumn{1}{c}{74.2} &\multicolumn{1}{c}{87.0} &\multicolumn{1}{c}{88.1} &\multicolumn{1}{c}{94.3} &\multicolumn{1}{c}{-} &\multicolumn{1}{c}{71.0} &\multicolumn{1}{c}{82.2} & \multicolumn{1}{c}{-} \\
    RoBERTa-large (FT) &355M &\multicolumn{1}{c}{86.0} &\multicolumn{1}{c}{87.6} &\multicolumn{1}{c}{92.0} &\multicolumn{1}{c}{96.5} &\multicolumn{1}{c}{-} &\multicolumn{1}{c}{68.5} &\multicolumn{1}{c}{78.2}& \multicolumn{1}{c}{-} \\
    \bottomrule
    \end{tabular}
}
    \caption{Comparison of our approach against many pre-trained open-source encoder-decoder and decoder-only Large Language Models on 7 Natural Language Understanding tasks in zero-shot conditions. FT stands for Full Finetuning and is included for reference. For \st, we report the average across 5 training runs and 5 evaluation runs and include the average standard deviation in parenthesis. We highlight all scores in gray where our approach with RoBERTa-base (best) exceeds or is equal to the score given by the model.}
    \label{tab:decoders-0shot}
\end{table*}

\section{Experimental Setup}


\subsection{Evaluation Datasets}

We measure our model's generalizability using another set of 7 diverse datasets representing a variety of unseen tasks or unseen domains: Balanced COPA (BCOPA; \citealp{kavumba-etal-2019-choosing, roemmele2011choice}), MRPC \cite{dolan-brockett-2005-automatically}, Emotion \cite{saravia-etal-2018-carer}, Amazon Polarity \cite{10.1145/2507157.2507163, zhangCharacterlevelConvolutionalNetworks2015}, FigQA \cite{https://doi.org/10.48550/arxiv.2204.12632}, StoryCloze (2016) \cite{yang2023improving}, and Yahoo Answers Topics \cite{zhangCharacterlevelConvolutionalNetworks2015}. Among the evaluation data, MRPC (paraphrase identification) and Amazon Polarity (sentiment analysis) represent tasks seen during training but in different domains and demonstrate \textit{cross-domain} generalizability. The rest are unseen tasks and hence examine the \textit{cross-task} generalizability.


\subsection{Statement Finetuning Configurations}

We statement-finetune both RoBERTa-base and RoBERTa-large across diverse NLP tasks outlined in Section~\ref{sec:st&a}. However, as statement fine-tuning expands the dataset with various templates over all possible labels, it is arguably unwise to fine-tune on all possible generated statements. Moreover, each task has a different data size, leading to unbalanced fine-tuning data. Therefore, we sample statements randomly for each task, uniformly across true and false statements. In true/false statements, we also balance original classes. We explore sample size from 1,000 statements to 50,000 statements per dataset. We encourage the invariance to phrasing in diverse statements by designing multiple statement templates per dataset (a list of all statement templates is shown in Appendix~\ref{sec:appendixa}). 
Furthermore, we 
run the training five times
to account for randomness in training data creation. In the evaluation, we randomly pick a template for each dataset in a single evaluation run and also repeat the evaluation five times. We thus report the mean and standard deviation of 5 $\times$ 5 runs to show the general task accuracy and the (in)variance to phrasing. We also explore the effect of statement diversity during training in Section~\ref{sec:spc}.

After multitask statement tuning is completed, we can further continue fine-tuning the model on the target downstream dataset. Specifically, we explore various n-shot configurations: Full/3,000-shot, 1,000-shot, 500-shot, 200-shot, and 32-shot, where we use limited data from the training sets of the corresponding dataset to fine-tune our statement-tuned models. For the Full/3,000-shot case, we cap the training set at 3,000 examples, otherwise, we use the entire set (this is the case for Amazon Polarity only). For StoryCloze, there is no training set, so we just carry out 32-shot (using 32 samples from the test set for fine-tuning and evaluating on the rest) and zero-shot experiments. As for Yahoo Answers Topic and Emotion, due to them being multi-class classification tasks, we cap the n-shot analysis at 200-shot due to the larger number of choices per example (and hence a larger number of statements per example).

\subsection{Other Baselines}

To assess the feasibility of our approach, we compare Statement-Tuned RoBERTa base models with 125 million parameters with a range of competitive multitask fine-tuned encoder-decoder models and decoder-only LLMs spanning a parameter range from 60 million parameters to 70 billion parameters. 
We include the following open-source models: Meta-Llama-3-70B-Instruct \cite{llama3modelcard}, Llama-2-13B-chat, Llama-2-7B-chat \cite{touvron2023llama}, Mistral-7B-Instruct-v0.2 \cite{jiang2023mistral}, QWEN1.5-7B-chat and QWEN1.5-0.5B-chat \cite{qwen}, Pythia-6.9B and Pythia-2.8B \cite{biderman2023pythia}, Phi-2 \cite{li2023textbooks}, FlanT5-Large and FlanT5-Small \cite{chung2022scaling}, and BART-large-mnli \cite{lewis2019bart}. 

We use the chat/instruction-tuned version of the models to allow for better instruction following. We try to select models that have not seen the evaluation data to the best of our knowledge, however, the training data of many of these models is not fully outlined and there can always be the possibility of contamination \cite{Li_Flanigan_2024}. Although these models have already been trained on a large number of instruction fine-tuning datasets, to guarantee a fair comparison as much as possible, we additionally instruction-tune a subset of the models using LoRA \cite{hu2021loralowrankadaptationlarge} on the same training datasets used for~\st.\footnote{Due to limited computational resources, we are only able to perform this extended training and analysis on a subset of the models ranging from 500M to 13B parameters.} Details regarding data formatting, hyper-parameters, and results are reported in Appendix~\ref{sec:appendixi}. 

We train and evaluate all the models on a configuration of 5 AMD EPYC Rome CPU cores and at most 4 Nvidia Tesla A100 40GB GPUs (we only use 4 GPUs for inference of the largest LLMs, and 1 GPU for \st).


\begin{figure*}[!h]
    \centering
    \includegraphics[width=0.85\linewidth]{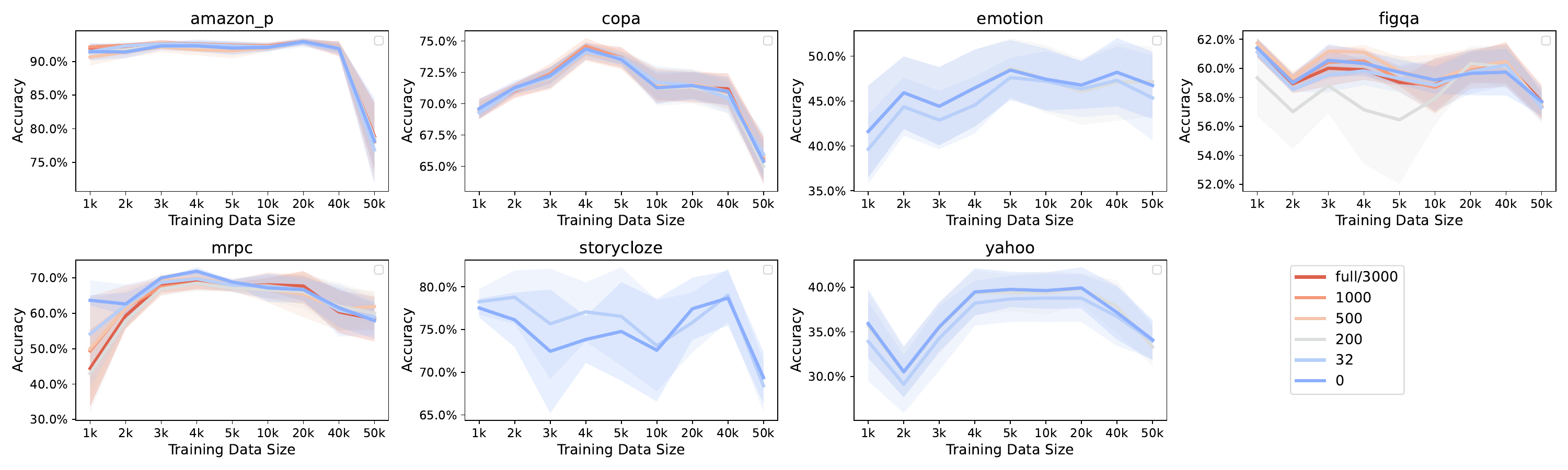}
    \caption{N-shot accuracy of Statement-Tuned RoBERTa-base models across training datasets of different sizes. The x-axis denotes the number of statements per \st~training dataset, with the number of training datasets fixed. 
    }
    \label{fig:model_size_vs_accuracy_n-shot}
\end{figure*}
\section{Results and Analysis}

In this section, we dive deep into the results of our experimentation to derive insights about our approach.

\subsection{Overall Result}\label{sec:5.1}

Table~\ref{tab:decoders-0shot} shows zero-shot performance of statement-tuned RoBERTa and baselines. Recall that we explore various statement-tuning sizes, hence here we report the best performance across all training sizes and performance for the 4,000 and 10,000 sample sizes per dataset for the base and large models, respectively. The effect of statement tuning sample size is explored in Section \ref{sec:sample_size}.

\rparagraph{Statement-Tuning Enables Effective Zero-Shot Generalization on Masked Language Models}
The result shows that the multitask statement-tuned encoder model can achieve zero-shot generalization across unseen tasks and domains. 
On BCOPA (unseen task) and Amazon Polarity (unseen domain), our zero-shot statement-tuned models even achieve accuracies on par with models fine-tuned on the full datasets.   
We also see that the larger model (RoBERTa-large) achieved much better generalization than the base model in general.

\rparagraph{Comparison Against Larger Zero-Shot Models}
Our approach is also competitive against other pre-trained open-source encoder-decoder and decoder-only Large Language Models under zero-shot prompting. Despite having significantly fewer parameters than all the models reported (except for FlanT5-small), our approach matches or exceeds many of them on the tasks reported. 
In average, our best Statement-Tuned RoBERTa-large models with only 355M parameters outperform the best performing LLM (Qwen1.5-7B-Chat) by 4.4 and the largest LLM (Meta-Llama-3-70B-Instruct, with approximately 200 times the number of parameters) by 4.8. 
It is worth noting that our RoBERTa-base models with only 125M parameters almost completely outperforms all models under or equal to 6.9B parameters (except for FlanT5-Large) on all tasks (except for BCOPA). Our models are dominant on FigQA and StoryCloze, both of which are unrepresented in the training data, 
with the best performing RoBERTa-large model scoring an additional \textbf{32.2} and \textbf{9.4} points over Llama3-70B-Instruct on the accuracy respectively.

We observe similar results in the 32-shot setting (see Appendix~\ref{sec:appendixc}) and when the LLMs are additionally instruction-tuned on the same data (see Appendix~\ref{sec:appendixi}). These results demonstrate the capabilities of much smaller encoder models as being accurate and light (in terms of parameters; for speed comparison see Appendix~\ref{sec:performance}) alternatives to LLM zero-shot (and few-shot) prompting in natural language understanding.


\subsection{Ablation Studies}

\begin{table}[!h]
\large
\centering
\resizebox{0.5\columnwidth}{!}{
\begin{tabular}{ccc} 
\toprule
\multirow{2}{*}{\begin{tabular}[c]{@{}c@{}}Statement\\Sample\end{tabular}} & \multicolumn{2}{c}{Average accuracy} \\ 
\cmidrule{2-3}
 & \multicolumn{1}{l}{RoB-base} & \multicolumn{1}{l}{RoB-large} \\ 
\midrule
1,000 & 63.0 & 71.1 \\
2,000 & 62.7 & 71.0 \\
3,000 & 63.1 & 73.3 \\
4,000 & \textbf{65.4} & 72.9 \\
5,000 & 64.7 & 72.2 \\
10,000 & 64.3 & \textbf{73.4} \\
20,000 & 64.9 & 68.6 \\
40,000 & 64.1 & 72.0 \\
50,000 & 58.5 & 68.2 \\
\bottomrule
\end{tabular}
}
\caption{Average accuracy over all evaluation tasks when trained with different statement sample size per dataset.}
\label{tab:statement-size}
\vspace{-0.5em}
\end{table}

\begin{figure*}[!h]
    \centering
    \includegraphics[width=0.85\linewidth]{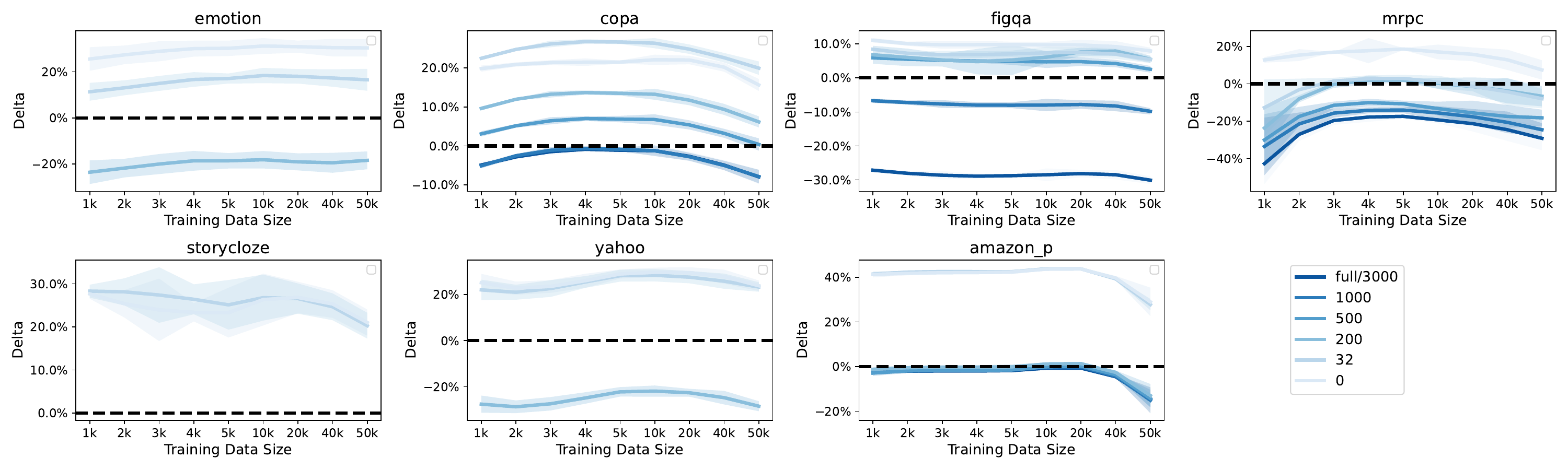}
    \caption{N-shot improvement of Statement-Tuned RoBERTa-base of varying training set sizes over standard fine-tuning. The y-axis, Delta, is the difference between the accuracy of the Statement-Tuned model and the accuracy achieved by regular fine-tuning of RoBERTa-base on the task. A positive Delta indicates improvement over the baseline approach.}
    \label{fig:model_size_vs_delta_n-shot}
\end{figure*}
\subsubsection{Statement Finetuning Sample Size}
\label{sec:sample_size}

Recall that we only perform statement fine-tuning on a sample of all possible statements from the training dataset. Here, we explore the effect of sample size per dataset in the multitask statement fine-tuning on both zero-shot and few-shot performance.

\rparagraph{Zero-Shot}
As shown in Table~\ref{tab:statement-size}, 
with only 1k samples per datasets, we can already reach 96\% of the best performance, which is obtained with 4k samples on RoBERTa-base and 10k samples on RoBERTa-large, showing the sample-efficiency of statement-tuning. 
For RoBERTa-base, introducing more data after 4,000 samples does not further improve the accuracy on downstream tasks and even causes a decrease. 
RoBERTa-large benefits from a larger fine-tuning data size with the best average performance observed when the statement number per training set increases to 10,000, more than doubling the optimal sample size of RoBERTA-base. We hypothesize that this is due to a larger capacity to understand and discriminate between natural language statements which allows RoBERTa-large to benefit from more training data as opposed to RoBERTa-base, which has a more limited capacity to develop a general semantic understanding of the truthfulness of statements. 
Nonetheless, we also observe a decrease in RoBERTa-large's performance after 10k samples. The existence of a point of diminishing returns in both models when it comes to \st~training data sizes indicates that 
too many samples may lead to overfitting to the training tasks, which affects the generalizability. This is consistent with previous observation that excessive instruction tuning can hurt the performance of smaller models (8B and below) on held-out tasks due to more limited model capacity \cite{DBLP:conf/iclr/WeiBZGYLDDL22}.

\rparagraph{Few-Shot} While the statement-tuned model shows zero-shot generalization, we can further fine-tune the model on the target downstream task. As seen in Figure~\ref{fig:model_size_vs_accuracy_n-shot}, we investigate the effect of both the multitask statement-tuning sample size and the number of shots from the target tasks on the n-shot performance on the 7 evaluation datasets. 

When increasing the multitask statement-tuning sample size, we observe a trend in n-shot performance similar to the general zero-shot performance shown in Table~\ref{tab:statement-size}. For example, the optimal data size is achieved at around 4k\textasciitilde5k on COPA, Emotion and Yahoo, and there is an apparent drop in accuracy across different shot numbers and tasks when increasing the sample size from 40k to 50k. There turns out to be a high degree of correlation among all n-shot and 0-shot performance (see Figure~\ref{fig:corr}), indicating that observed trends in the 0-shot scenario can be informative for the few-shot cases.  



However, the results seem to indicate a general trend of diminishing returns past using 200-shot finetuning. Nevertheless, it seems that a great deal of the potential performance is achieved with the zero-shot application of the approach, hence further supporting the utility of our approach when task-specific data is scarce. 

\subsubsection{Comparison with Standard Fine-Tuning}

To observe the improvement over regular fine-tuning of RoBERTa-base, we also include Figure~\ref{fig:model_size_vs_delta_n-shot}, where the y-axis, Delta, represents the improvement over regular fine-tuning for the particular n-shot. For zero-shot, we take random choice as the baseline. Generally, continually fine-tuning our model is better than fine-tuning vanilla RoBERTa under an extremely low N-shot setting. However, in some instances such as BCOPA and (to a certain extent) FigQA, we notice that even for a higher number of few-shot examples, we tend to observe a benefit against regular fine-tuning. 

Our approach is recommended in extreme few-shot and zero-shot scenarios. When more data is available, directly fine-tuning RoBERTa-base is better. Our method’s good performance with limited data can be attributed to improved generalizability from multitask statement tuning and data augmentation effect of statements generated from few-shot examples, which enhances data efficiency.


\begin{table}[t]\centering\small
\resizebox{\columnwidth}{!}{
\begin{tabular}{@{}c@{ ~ }c@{ ~ }c@{ ~ }c@{ ~ }c@{ ~ }c@{ ~ }c@{ ~ }c|c@{}}
\toprule
\textbf{SPC} &\textbf{BCOPA} &\textbf{MRPC} &\textbf{FigQA} &\textbf{AP} &\textbf{S-Cloze} &\textbf{YA Topic} &\textbf{Emotion} &\textbf{AVG} \\
\midrule
1 &91.0$_{(1.7)}$ &\textbf{74.8}$_{(2.2)}$ &\textbf{49.8}$_{(4.8)}$ &59.1$_{(0.5)}$ &58.4$_{(17.1)}$ &78.0$_{(3.4)}$ &\textbf{41.2}$_{(2.1)}$ &\textbf{62.6}$_{(6.9)}$ \\
2 &91.3$_{(1.0)}$ &70.2$_{(1.4)}$ &49.2$_{(3.1)}$ &\textbf{61.6 }$_{(1.1)}$&56.9$_{(8.3)}$ &\textbf{79.9}$_{(2.2)}$ &33.5$_{(\textbf{1.4})}$ &60.5$_{(3.6)}$ \\
3 &\textbf{93.0}$_{(\textbf{0.3})}$ &73.2$_{(\textbf{0.7})}$ &43.2$_{(\textbf{1.0})}$ &60.5$_{(1.1)}$ &64.3$_{(6.7)}$ &74.2$_{(3.0)}$ &31.3$_{(2.4)}$ &59.5$_{(\textbf{3.0})}$ \\
4 &92.1$_{(\textbf{0.3})}$ &70.9$_{(1.6)}$ &49.4$_{(\textbf{1.0})}$ &59.9$_{(\textbf{0.4})}$ &\textbf{68.1}$_{(\textbf{6.4})}$ &68.0$_{(7.5)}$ &38.8$_{(2.1)}$ &61.9$_{(3.9)}$ \\
5 &92.4$_{(0.5)}$ &69.6$_{(1.0)}$&46.2$_{(3.5)}$ &60.5$_{(1.1)}$ &66.8$_{(6.7)}$ &78.5$_{(\textbf{2.1})}$ &35.9$_{(3.0)}$ &61.6$_{(3.2)}$ \\
\bottomrule
\end{tabular}
}
\caption{The Zero-shot performance of the base model 
using various degrees of SPC,
where a larger SPC value indicates greater statement diversity during training. We report the average as the geometric mean of the task performance to account for the differing accuracy ranges of each task. Each value is a mean over 5 evaluation runs and we include the standard deviation in the parentheses.}\label{tab:SPC}
\vspace{-1.0em}
\end{table}

\begin{table*}[]
    \centering
    \resizebox{\textwidth}{!}{\begin{tabular}{ccccccccc|ccccccc:c}
    \toprule
        \multicolumn{9}{c}{\textbf{Statement tuning Training}} & \multicolumn{7}{c}{\textbf{Evaluation}} \\
         \textbf{PD} &\textbf{CR} &\textbf{NLI} &\textbf{QnA} &\textbf{SA} &\textbf{WSD} &\textbf{IC} &\textbf{OLI} &\textbf{SU} &\textbf{BCOPA} &\textbf{MRPC} &\textbf{FIGQA} &\textbf{AMAZON P.} &\textbf{StoryCloze} &\textbf{YA Topic} &\textbf{Emotion} &\textbf{AVG} \\\midrule
        x &x &x &x &x &x &x &x &x &71.0$_{(0.9)}$ & 65.7$_{(3.1)}$ & 59.8$_{(1.0)}$ & 90.7$_{(1.3)}$ & 75.1$_{(3.8)}$ & 36.8$_{(2.7)}$ & 46.2$_{(3.8)}$ & 61.2$_{(2.7)}$ \\
        x &x &x &x &x &x &x & x & &69.8$_{(2.6)}$ & 65.9$_{(6.1)}$ & 60.4$_{(0.4)}$ & 91.2$_{(0.7)}$ & 79.8$_{(1.7)}$ & 29.4$_{(3.7)}$ & 47.1$_{(0.4)}$ & 60.0$_{(3.0)}$ \\
        x &x &x &x &x &x &x & & &70.0$_{(0.3)}$ & 64.2$_{(7.9)}$ & 59.3$_{(0.2)}$ & 92.0$_{(0.3)}$ & 70.5$_{(6.4)}$ & 31.2$_{(2.3)}$ & 49.4$_{(3.0)}$ & 59.6$_{(4.1)}$ \\
        x &x &x &x &x &x & & & &68.7$_{(2.1)}$ & 64.6$_{(6.8)}$ & 58.8$_{(0.7)}$ & 91.3$_{(0.5)}$ & 77.3$_{(7.0)}$ & 18.7$_{(3.0)}$ & 53.4$_{(4.9)}$& 56.5$_{(4.4)}$ \\
        x &x &x &x &x & & & & &70.2$_{(0.5)}$ & 67.0$_{(4.6)}$ & 59.4$_{(0.8)}$ & 91.8$_{(0.1)}$ & 73.4$_{(6.7)}$ & 20.5$_{(3.4)}$ & 52.2$_{(3.0)}$ & 57.2$_{(3.5)}$ \\
        x &x &x &x & & & & & &70.0$_{(1.2)}$ &67.4$_{(2.5)}$ &59.2$_{(0.3)}$ &78.0$_{(12.0)}$ &75.3$_{(10.1)}$ &36.6$_{(3.3)}$ &40.3$_{(2.2)}$ &58.8$_{(6.2)}$ \\
        x &x &x & & & & & & &50.2$_{(1.7)}$ & 40.6$_{(8.5)}$ & 50.9$_{(1.7)}$ & 55.9$_{(5.5)}$ & 50.2$_{(6.6)}$ & 3.8$_{(1.8)}$ & 7.2$_{(2.8)}$ & 26.0$_{(4.8)}$ \\
    \bottomrule
    \end{tabular}}
    \caption{Comparison of the effect of reducing task diversity in the training of \st~models on zero-shot accuracy on unseen datasets. The last column is the average using the geometric mean to account for the different accuracy ranges of the different evaluation sets. The total training set size remains constant at approximately 100,000 statements across all configurations.}
    \label{tab:diversity}
\end{table*}

\subsubsection{Effect of Statement Diversity}
\label{sec:spc}

As part of our investigation of Statement-Tuning, we would like to explore the effect of template diversity during Statement-Tuning. We hypothesize that randomly applying a larger number of different statement templates per training corpus will allow for improved performance on unseen tasks, as it will make the model more robust to the phrasing of statement templates and prevent it from relying on superficial cues in certain templates.

In our main experiments, each dataset employs several templates (see Appendix~\ref{sec:appendixa}). In this experiment, we limit each corpus to only use the maximum of N different templates, which we call Statements per Category (SPC). We statement-tune RoBERTa base models with a fixed training set size of 4,000 statements per training corpus with a varying level of SPC.

Table~\ref{tab:SPC} shows that though BCOPA and StoryCloze benefits from a larger SPC, increasing SPC doesn’t always boost average task performance, with the highest being 62.6 at SPC 1. However, average standard deviation drops significantly from 6.9\% to 3.6\% when SPC increases from 1 to 2, reaching a low of 3.0\% at SPC 3. This suggests that more template diversity improves stability and consistency. Therefore, using at least 2 different templates is recommended for better robustness. 

\subsubsection{Effect of Task Diversity}

We examine the importance of task variety in \st. Our \st~datasets can be grouped into 9 task categories: Summarization (SU), Sentiment Analysis (SA), Question Answering (QA), Natural Language Inference (NLI), Commonsense Reasoning (CR), Paraphrase Detection (PD), Word Sense Disambiguation (WSD), Intent Classification (IC), and Offensive Language Identification (OLI). See Appendix~\ref{sec:appendixe} for the dataset breakdown. We perform statement tuning on RoBERTa-base with various task subsets, dynamically sampling data to maintain 100k total statements.



Table~\ref{tab:diversity} shows the zero-shot performance of the statement tuning approach with a fixed training set size but varying task types. Average performance increases from 26.0 to 61.2 as the number of tasks is increased from the minimum of 3 to the maximum of 9. Robustness also improves, shown by a decrease in average standard deviation from 4.8\% to 2.7\%. This demonstrates that increasing training task diversity can boost performance and reduce variance.
Unsurprisingly, the inclusion of the Sentiment Analysis task substantially improves   
the performance on Amazon Polarity from the same task category. Another related task, Emotion, also shows a large increase after adding the SA task. Although the inclusion of SA and WSD hurts the performance on a dissimilar task, Yahoo Answer Topic, the accuracy is recovered after adding the more related Intent Classification task, and reaches the highest 36.8 when training with all tasks. 
However, the enhancement of downstream tasks does not always come from similar training tasks. More interestingly, adding the QA task leads to a significant jump in the performance of all evaluation tasks. Though Paraphrase Identification is always included in the training, MRPC still benefits from the QA task, reflected by a great improvement of 26.8. Both related and unrelated training tasks can have a positive effect on the downstream tasks, highlighting the value of task diversity in multitask statement-tuning. Sometimes adding an unrelated task causes a performance drop on certain datasets, e.g. StoryCloze after adding IC, but including more tasks alleviates the problem, again confirming the advantage of task diversity.

\section{Conclusion}
As part of their emergent abilities, LLMs generalize to many unseen tasks/domains through few-shot and zero-shot prompting, but are prohibitively computationally expensive and difficult to adapt. To address this issue, we investigate \st, a novel technique for few-shot and zero-shot task generalization for encoder models. We find that this approach can match/outperform the few-shot and zero-shot prompting of many much larger decoder-only or encoder-decoder models on many tasks at a fraction of the parameters. Experimentation shows that the approach can be leveraged by training on as few as 16,000 statements. We find training task and statement template diversity to be generally helpful. We speculate that the benefits of this approach could extend beyond task generalization and could prove useful for cross-lingual task transfer, and would like to explore this in future work.
\section*{Limitations}
While our approach offers advantages in computational efficiency compared to LLMs, the cost scales with the number of possible targets due to the requirement of one forward pass per label. That being said, it is still possible to apply our method in extreme multi-class classification because we do not have to use all possible statements with all possible labels for training, as our model is trained exactly to generalize to unseen labels by learning the relation of label semantics and the input text.
Additionally, task-specific \textit{full} fine-tuning can still achieve better performance in the presence of more training data. Therefore, we recommend the use of our approach in the low-resource/no-resource scenario. 
Furthermore, our method can be sensitive to the \st~training set size and other hyperparameters, hence some exploration of ideal hyperparameters may be required before employing Statement-Tuned models. In addition, we limit our analysis only to English, it would be interesting to observe whether the technique enables cross-lingual transfer but we leave this to future work. Finally, our reliance on encoder-based models restricts its application to Natural Language Understanding tasks, excluding tasks like translation or abstractive summarization.

\section*{Ethics Statement}
We affirm our commitment to more accessible and climate-aware NLP, and hope this work inspires more computationally efficient approaches to NLP. All data and models we use are publicly available. Furthermore, the success of \st~relies on fine-tuning pretrained encoder models, which are pretrained on large datasets, and hence, \st~is susceptible to inheriting and enforcing any harmful biases existing in the pretraining data.

\section*{Acknowledgements}
Yongxin Huang is supported by HUAWEI Technologies (Ireland) Co., Ltd.
\bibliography{anthology,custom}
\bibliographystyle{acl_natbib}
\clearpage

\appendix
\section{Statement Templates}
\label{sec:appendixa}
\subsection{QQP Templates}

\resizebox{\columnwidth}{!}{
    \begin{tabular}{ll}\toprule
    \textbf{Task} &\textbf{Statement Template} \\\midrule
    \multirow{4}{*}{QQP} &"\{\{text1\}\}" is a duplicate of "\{\{text2\}\}" \\
    &"\{\{text1\}\}" duplicates \{\{text2\}\} \\
    &"\{\{text1\}\}" is not a duplicate of "\{\{text2\}\}" \\
    &"\{\{text1\}\}" does not duplicate "\{\{text2\}\}" \\
    \bottomrule
    \end{tabular}
}

\subsection{Winogrande Templates}

\resizebox{\columnwidth}{!}{
\begin{tabular}{ll}\toprule
\textbf{Task} &\textbf{Statement Template} \\\midrule
\multirow{5}{*}{Winogrande} &In "\{\{sentence\}\}", \_ is: \{\{option1/option2\}\} \\
&Q: "\{\{sentence\}\}", A: \{\{option1/option2\}\} \\
&The missing word in: "\{\{sentence\}\}" is \{\{option1/option2\}\} \\
&\_ in: "\{\{sentence\}\}" is \{\{option1/option2\}\} \\
&"\{\{sentence\}\}", \_ is: \{\{option1/option2\}\} \\
\bottomrule
\end{tabular}
}

\subsection{PiQA Templates}

\resizebox{\columnwidth}{!}{
\begin{tabular}{ll}\toprule
\textbf{Task} &\textbf{Statement Template} \\\midrule

\multirow{4}{*}{PiQA} &\{\{goal\}\} \{\{sol1/sol2\}\} \\
&Goal: \{\{goal\}\}, Solution: \{\{sol1/sol2\}\} \\
&If the goal is: \{\{goal\}\}, then the solution is: \{\{sol1/sol2\}\} \\
&Problem: \{\{goal\}\}, Solution: \{\{sol1/sol2\}\} \\
\bottomrule
\end{tabular}
}

\subsection{MNLI and SNLI Templates}

\resizebox{\columnwidth}{!}{
\begin{tabular}{ll}\toprule
\textbf{Task} &\textbf{Statement Template} \\\midrule
\multirow{9}{*}{MNLI} &"\{\{text1\}\}" entails "\{\{text2\}\}" \\
&\{\{text1\}\}? yes, \{\{text2\}\} \\
&Premise: \{\{text1\}\}, Hypothesis: \{\{text2\}\}, label: Entailment \\
&"\{\{text1\}\}" is neutral with regards to "\{\{text2\}\}" \\
&\{\{text1\}\}? maybe, \{\{text2\}\} \\
&Premise: \{\{text1\}\}, Hypothesis: \{\{text\}\}, label: Neutral \\
&"\{\{text1\}\}" contradicts "\{\{text2\}\}" \\
&\{\{text1\}\}? no, \{\{text2\}\} \\
&Premise: \{\{text1\}\}, Hypothesis: \{\{text\}\}, label: Contradiction \\
\bottomrule
\end{tabular}
}

\subsection{Mintaka Templates}

\resizebox{\columnwidth}{!}{
\begin{tabular}{ll}\toprule
\textbf{Task} &\textbf{Statement Template} \\\midrule
\multirow{4}{*}{Mintaka} &Q: \{\{question\}\}, A: \{\{answerText\}\} \\
&\{\{question\}\} \{\{answerText\}\} \\
&Question: \{\{question\}\}, Answer: \{\{answerText\}\} \\
&The answer of \{\{question\}\} is \{\{answerText\}\} \\
\bottomrule
\end{tabular}
}

\subsection{Yelp Polarity Templates}

\resizebox{\columnwidth}{!}{
\begin{tabular}{ll}\toprule
\textbf{Task} &\textbf{Statement Template} \\\midrule
\multirow{10}{*}{Yelp Polarity} &"Title: \{\{title\}\}, Content: \{\{content\}\}" has negative sentiment \\
&\{\{title\}\} \{\{content\}\} has negative sentiment \\
&"Title: \{\{title\}\}, Content: \{\{content\}\}", Sentiment: Negative \\
&\{\{title\}\} \{\{content\}\} It was terrible \\
&The sentiment in "\{\{title\}\} \{\{content\}\}" is negative \\
&"Title: \{\{title\}\}, Content: \{\{content\}\}" has positive sentiment \\
&\{\{title\}\} \{\{content\}\} has positive sentiment \\
&"Title: \{\{title\}\}, Content: \{\{content\}\}", Sentiment: Positive \\
&\{\{title\}\} \{\{content\}\} It was great \\
&The sentiment in "\{\{title\}\} \{\{content\}\}" is positive \\
\bottomrule
\end{tabular}
}

\subsection{WikiLingua Templates}

\resizebox{\columnwidth}{!}{
\begin{tabular}{ll}\toprule
\textbf{Task} &\textbf{Statement Template} \\\midrule
\multirow{5}{*}{WikiLingua} &Passage: \{\{source\}\}, Summary: \{\{target\}\} \\
&The summary of "\{\{source\}\}" is \{\{target\}\} \\
&Context: \{\{source\}\}, Summary: \{\{target\}\} \\
&Q: Summarize the following: \{\{source\}\}, A: \{\{target\}\} \\
&The answer of "Summarize the following \{\{source\}\}" is \{\{target\}\} \\
\bottomrule
\end{tabular}
}

\subsection{SQuAD Templates}
\resizebox{\columnwidth}{!}{
\begin{tabular}{ll}\toprule
\textbf{Task} &\textbf{Statement Template} \\\midrule
\multirow{4}{*}{SQuAD} &Context: \{\{context\}\}\textbackslash n Question: \{\{question\}\}\textbackslash n Answer: \{\{answers/random\_span\}\} \\
&\{\{context\}\}\textbackslash n According to the passage above, the answer of \{\{question\}\} is \{\{answers/random\_span\}\} \\
&"Passage: \{\{context\}\}\textbackslash n Question: \{\{question\}\}\textbackslash n Answer: \{\{answers/random\_span\}\} \\
&\{\{context\}\}\textbackslash n Q: \{\{question\}\}\textbackslash n A:\{\{answers/random\_span\}\} \\
\bottomrule
\end{tabular}
}

\subsection{BCOPA Templates}

\resizebox{\columnwidth}{!}{
\begin{tabular}{ll}\toprule
\textbf{Task} &\textbf{Statement Template} \\\midrule
\multirow{6}{*}{BCOPA} &The cause of \{\{premise\}\} is that \{\{choice1/choice2\}\} \\
&\{\{premise\}\} because \{\{choice1/choice2\}\} \\
&\{\{premise\}\} due to \{\{choice1/choice2\}\} \\
&The effect of \{\{premise\}\} is that \{\{choice1/choice2\}\} \\
&\{\{premise\}\} therefore \{\{choice1/choice2\}\} \\
&\{\{premise\}\}, so \{\{choice1/choice2\}\} \\
\bottomrule
\end{tabular}
}

\subsection{MRPC Templates}

\resizebox{\columnwidth}{!}{
\begin{tabular}{ll}\toprule
\textbf{Task} &\textbf{Statement Template} \\\midrule
\multirow{5}{*}{MRPC} &"\{\{text1\}\}" is a paraphrase of "\{\{text2\}\}" \\
&"\{\{text1\}\}"\textbackslash n In other words: "\{\{text2\}\}" \\
&\{\{text1\}\}? yes, \{\{text2\}\} \\
&"\{\{text1\}\}" can be stated as "\{\{text2\}\}" \\
&\{\{text1\}\}" is the same as saying "\{\{text2\}\}" \\
\bottomrule
\end{tabular}
}

\subsection{Amazon Polarity Templates}

\resizebox{\columnwidth}{!}{
\begin{tabular}{ll}\toprule
\textbf{Task} &\textbf{Statement Template} \\\midrule
\multirow{12}{*}{Amazon Polarity} &"Title: \{\{title\}\}, Content: \{\{content\}\}" has negative sentiment \\
&\{\{title\}\} \{\{content\}\} has negative sentiment \\
&"Title: \{\{title\}\}, Content: \{\{content\}\}", Sentiment: Negative \\
&\{\{title\}\} \{\{content\}\} It was terrible \\
&The sentiment in "\{\{title\}\} \{\{content\}\}" is negative \\
&The emotions conveyed in "\{\{title\}\} \{\{content\}\}" are negative \\
&"Title: \{\{title\}\}, Content: \{\{content\}\}" has positive sentiment \\
&\{\{title\}\} \{\{content\}\} has positive sentiment \\
&"Title: \{\{title\}\}, Content: \{\{content\}\}", Sentiment: Positive \\
&\{\{title\}\} \{\{content\}\} It was great \\
&The sentiment in "\{\{title\}\} \{\{content\}\}" is positive \\
&The emotions conveyed in "\{\{title\}\} \{\{content\}\}" are positive \\
\bottomrule
\end{tabular}
}

\subsection{FigQA Templates}

\resizebox{\columnwidth}{!}{
\begin{tabular}{ll}\toprule
\textbf{Task} &\textbf{Statement Template} \\\midrule
\multirow{5}{*}{FigQA} &\{\{startphrase\}\} \{\{ending1/ending2\}\} \\
&\{\{startphrase\}\} therefore \{\{ending1/ending2\}\} \\
&startphrase: \{\{startphrase\}\}, ending: \{\{ending1/ending2\}\} \\
&if \{\{startphrase\}\} then \{\{ending1/ending2\}\} \\
&\{\{startphrase\}\} means \{\{ending1/ending2\}\} \\
\bottomrule
\end{tabular}
}

\subsection{StoryCloze Templates}

\resizebox{\columnwidth}{!}{
\begin{tabular}{ll}\toprule
\textbf{Task} &\textbf{Statement Template} \\\midrule
\multirow{3}{*}{StoryCloze} &\{\{input\_sentence\_1\}\} \{\{input\_sentence\_2\}\} \\ & \{\{input\_sentence\_3\}\} \{\{input\_sentence\_4\}\} \\ & \{\{sentence\_quiz1/sentence\_quiz2\}\} \\
\bottomrule
\end{tabular}
}

\subsection{Yahoo Topics Answers Templates}

\resizebox{\columnwidth}{!}{
\begin{tabular}{ll}\toprule
\textbf{Task} &\textbf{Statement Template} \\\midrule
YA Topic &\{\{question\_title\}\} \{\{question\_content\}\} the topic is \{\{topic\}\} \\

\bottomrule
\end{tabular}
}

\subsection{Emotion Templates}

\resizebox{\columnwidth}{!}{
\begin{tabular}{ll}\toprule
\textbf{Task} &\textbf{Statement Template} \\\midrule
Emotion &\{\{question\_title\}\} \{\{question\_content\}\} the topic is \{\{topic\}\} \\
\bottomrule
\end{tabular}
}

\subsection{Offensive Templates}

\resizebox{\columnwidth}{!}{
\begin{tabular}{ll}\toprule
\textbf{Task} &\textbf{Statement Template} \\\midrule
\multirow{5}{*}{Offensive} &"\{\{text\}\}". The tweet is \{\{label\}\}. \\
&This tweet "\{\{text\}\}" is considered \{\{label\}\}. \\
&Tweet: "\{\{text\}\}". Label: \{\{label\}\}. \\
&"\{\{text\}\}". This text is \{\{label\}\}. \\
&The text "\{\{text\}\}" is \{\{label\}\}. \\
\bottomrule
\end{tabular}
}

\subsection{Massive Templates}

\resizebox{\columnwidth}{!}{
\begin{tabular}{ll}\toprule
\textbf{Task} &\textbf{Statement Template} \\\midrule
\multirow{4}{*}{Massive} &The utterance "\{\{utt\}\}" is under the \{\{scenario\}\} scenario. \\
&Utterance: "\{\{utt\}\}" Scenario: \{\{scenario\}\} \\
&User: "\{\{utt\}\}". The best scenario for the user query is \{\{scenario\}\}. \\
&The scenario of user's utterance "\{\{utt\}\}" is \{\{scenario\}\}. \\
\bottomrule
\end{tabular}
}

\subsection{Definite Pronoun Resolution Templates}
\resizebox{\columnwidth}{!}{
\begin{tabular}{ll}\toprule
\textbf{Task} &\textbf{Statement Template} \\\midrule
\multirow{4}{*}{DPR} &\{\{sentence\_with\_pronoun\_replaced\}\} \\
&\{\{sentence\}\}. Based on the sentence, \{\{pronoun\}\} refers to \{\{candidates\}\}. \\
&The pronoun \{\{pronoun\}\} in "\{\{sentence\}\}" is referring to \{\{candidates\}\}. \\
&\{\{sentence\}\}. '\{\{pronoun\}\}' refers to \{\{candidates\}\}. \\
\bottomrule
\end{tabular}
}

\subsection{QASC Templates}

\resizebox{\columnwidth}{!}{
\begin{tabular}{ll}\toprule
\textbf{Task} &\textbf{Statement Template} \\\midrule
\multirow{8}{*}{QASC} &\{\{formatted\_question\}\}. Answer: \{\{answer\_key\}\} \\
&Q: "\{\{formatted\_question\}\}." A: \{\{answer\_key\}\} \\
&Question: "\{\{formatted\_question\}\}." Answer: \{\{choices[answer\_key]\}\} \\
&Context: \{\{combined\_facts\}\} Question: \{\{question\}\} Answer: \{\{choices[answer\_key]\}\} \\
&\{\{question\}\} Based on the passage "\{\{combined\_facts\}\}", the answer if the question is "\{\{choices[answer\_key]\}\}". \\
&\{\{combined\_facts\}\} \{\{question\}\} \{\{choices[answer\_key]\}\} \\
&Context: \{\{combined\_facts\}\} Question: \{\{formatted\_question\}\}. Answer: \{\{answer\_key\}\} \\
&\{\{formatted\_question\}\}. The answer is \{\{answer\_key\}\} \\
\bottomrule
\end{tabular}
}

\subsection{SciQ Templates}

\resizebox{\columnwidth}{!}{
\begin{tabular}{ll}\toprule
\textbf{Task} &\textbf{Statement Template} \\\midrule
\multirow{5}{*}{SciQ} &\{\{question\}\} \{\{correct\_answer\}\} \\
&Question: \{\{question\}\} Answer: \{\{correct\_answer\}\} \\
&\{\{support\}\} Question: \{\{question\}\} Answer: \{\{correct\_answer\}\} \\
&\{\{support\}\} According to the information, \{\{question\}\}. Answer: \{\{correct\_answer\}\}. \\
&The answer to the question \{\{question\}\}, according to "\{\{support\}\}" is \{\{correct\_answer\}\}. \\
\bottomrule
\end{tabular}
}

\subsection{RACE Templates}

\resizebox{\columnwidth}{!}{
\begin{tabular}{ll}\toprule
\textbf{Task} &\textbf{Statement Template} \\\midrule
RACE &\{\{article\}\} \{\{question\_replaced\_with\_answer\}\} \\
\bottomrule
\end{tabular}
}

\clearpage

\section{Finetuning Setup}
\label{sec:appendixb}
To finetune RoBERTa-base/RoBERTa-large on \st, we train for 15 epochs using an initial learning rate of 1e-06 and a weight decay of 0.01. We use a warm-up ratio of 0.1. We use 10\% of the training data for validation. We use a training batch size of 16 for RoBERTa-base and a training batch size of 8 for RoBERTa-large.

\section{32-shot Generalization}
Table \ref{tab:decoders-32shot} shows the results of 32-shot fine-tuning on 7 target downstream datasets of our models and baselines. We observe similar trends to zero-shot setting as discussed in Section \ref{sec:5.1}.

\label{sec:appendixc}
\begin{table*}[!h]
    \centering
    \resizebox{\textwidth}{!}{\begin{tabular}[width=\linewidth]{lcccccccc|c}
    \hline
         &  \textbf{\#Parameters} &   \multicolumn{1}{c}{\textbf{BCOPA}}   &   \multicolumn{1}{c}{\textbf{MRPC}}   &   \multicolumn{1}{c}{\textbf{FigQA}}   &   \multicolumn{1}{c}{\textbf{Amazon Polarity}}   &   \multicolumn{1}{c}{\textbf{StoryCloze}}   &   \multicolumn{1}{c}{\textbf{Yahoo Answers Topic}}   &   \multicolumn{1}{c}{\textbf{Emotion}} &  \multicolumn{1}{c}{\textbf{Avg}} \\
         \hline
         Meta-Llama-3-70B-Instruct &70B &95.2 &78.9 &\textcolor{lightgray}{\textbf{46.1}} &96.6 &86.5 &66.4 &58.8 & 75.5 \\
Llama-2-13b-chat-hf &13B &93.2 &71.6 &\textcolor{lightgray}{\textbf{44.3}} &95.7 &84.9 &61.1 &56.6 & 72.5 \\
         Llama-2-7b-chat &7B &91.0 &\textcolor{lightgray}{\textbf{67.9}} &\textcolor{lightgray}{\textbf{42.8}} &95.2 &82.1 &61.9 &54.3 & 70.7 \\
Mistral-7B-Instruct-v0.2 &7B &93.8 &78.2 &\textcolor{lightgray}{\textbf{44.8}} &96.2 &87.0 &65.0 &57.0 & 74.6 \\
Qwen1.5-7B-Chat &7B &91.4 &79.4 &\textcolor{lightgray}{\textbf{43.8}} &95.1 &82.4 &63.9 &58.0 & 73.4 \\
Pythia-6.7B &6.7B &84.6 &\textcolor{lightgray}{\textbf{66.9}} &\textcolor{lightgray}{\textbf{39.2}} &\textcolor{lightgray}{\textbf{91.6}} &\textcolor{lightgray}{\textbf{74.0}} &\textcolor{lightgray}{\textbf{38.3}} &52.0 & 63.8 \\
Pythia-2.7B &2.7B &80.8 &\textcolor{lightgray}{\textbf{63.5}} &\textcolor{lightgray}{\textbf{41.5}} &\textcolor{lightgray}{\textbf{90.8}} &\textcolor{lightgray}{\textbf{71.7}} &\textcolor{lightgray}{\textbf{35.5}} &\textcolor{lightgray}{\textbf{47.5}} & 61.6 \\
Phi-2 &2.7B &90.8 &74.0 &\textcolor{lightgray}{\textbf{44.7}} &93.8 &\textcolor{lightgray}{\textbf{81.6}} &58.4 &58.4 & 71.7 \\
FlanT5-Large &770M &\textcolor{lightgray}{\textbf{66.2}} &78.7 &\textcolor{lightgray}{\textbf{39.7}} &\textcolor{lightgray}{\textbf{75.3}} &\textcolor{lightgray}{\textbf{59.9}} &\textcolor{lightgray}{\textbf{38.0}} &\textcolor{lightgray}{\textbf{34.6}} & 56.1\\
Qwen1.5-0.5B-Chat &500M &\textcolor{lightgray}{\textbf{73.4}} &\textcolor{lightgray}{\textbf{56.1}} &\textcolor{lightgray}{\textbf{38.5}} &\textcolor{lightgray}{\textbf{84.2}} &\textcolor{lightgray}{\textbf{68.8}} &\textcolor{lightgray}{\textbf{36.1}} &\textcolor{lightgray}{\textbf{31.4}} & 55.5\\
BART-large-mnli &406M &\textcolor{lightgray}{\textbf{52.2}} &\textcolor{lightgray}{\textbf{32.4}} &\textcolor{lightgray}{\textbf{42.0}} &\textcolor{lightgray}{\textbf{50.6}} &\textcolor{lightgray}{\textbf{51.1}} &\textcolor{lightgray}{\textbf{7.1}} &\textcolor{lightgray}{\textbf{10.0}} & 35.0\\
FlanT5-Small &60M &\textcolor{lightgray}{\textbf{52.0}} &\textcolor{lightgray}{\textbf{32.6}} &\textcolor{lightgray}{\textbf{41.4}} &\textcolor{lightgray}{\textbf{75.8}} &\textcolor{lightgray}{\textbf{50.0}} &\textcolor{lightgray}{\textbf{9.1}} &\textcolor{lightgray}{\textbf{9.8}} & 38.7\\
\hdashline
\textbf{Our Approach: RoBERTa-base (Best)}&\textbf{125M} &75.0$_{(0.5)}$ &70.0$_{(15.2)}$ &61.1$_{(0.4)}$ &92.8$_{(1.0)}$ &79.7$_{(1.5)}$ &39.2$_{(4.4)}$ &48.1$_{(3.9)}$ & 66.6 \\
\textbf{Our Approach: RoBERTa-base (4k)} &\textbf{125M} &75.0$_{(0.6)}$ &70.0$_{(1.9)}$ &60.3$_{(1.0)}$ &92.4$_{(0.8)}$ &79.7$_{(3.5)}$ &38.2$_{(2.5)}$ &45.4$_{(3.2)}$ & 65.9 \\
\textbf{Our Approach: RoBERTa-large (Best)} &\textbf{355M} &85.1$_{(0.8)}$ &71.5$_{(1.9)}$ &74.7$_{(1.8)}$ &95.3$_{(0.8)}$ &91.9$_{(0.2)}$ &50.2$_{(2.2)}$ &49.8$_{(1.4)}$ & 74.1 \\
\textbf{Our Approach: RoBERTa-large (10k)} &\textbf{355M} & 85.1$_{(0.8)}$ &71.5$_{(0.8)}$ &72.6$_{(1.8)}$ &95.3$_{(0.3)}$ &91.0$_{(1.0)}$ &48.2$_{(0.8)}$ &48.4$_{(3.6)}$ & 73.1 \\
    \hline
    \multicolumn{9}{l}{\textbf{Full-shot:}}\\
    RoBERTa-base (FT) &125M &\multicolumn{1}{c}{74.2} &\multicolumn{1}{c}{87.0} &\multicolumn{1}{c}{88.1} &\multicolumn{1}{c}{94.3} &\multicolumn{1}{c}{-} &\multicolumn{1}{c}{71.0} &\multicolumn{1}{c}{82.2} & \multicolumn{1}{c}{-} \\
    RoBERTa-large (FT) &355M &\multicolumn{1}{c}{86.0} &\multicolumn{1}{c}{87.6} &\multicolumn{1}{c}{92.0} &\multicolumn{1}{c}{96.5} &\multicolumn{1}{c}{-} &\multicolumn{1}{c}{68.5} &\multicolumn{1}{c}{78.2} & \multicolumn{1}{c}{-} \\
    \end{tabular}
    }
    \caption{Comparison of our approach against many pretrained open-source encoder-Decoder and Decoder-only Pretrained Large Language Models on 7 Natural Language Understanding tasks in 32-shot conditions. We highlight all scores in gray where our approach with RoBERTa-base (best) exceeds or is equal to the score given by the model.}
    \label{tab:decoders-32shot}
\end{table*}

\section{Regular Classification of Statement-Tuned Models}
\label{sec:appendixd}
\begin{figure*}
    \centering
    \includegraphics[width=\linewidth]{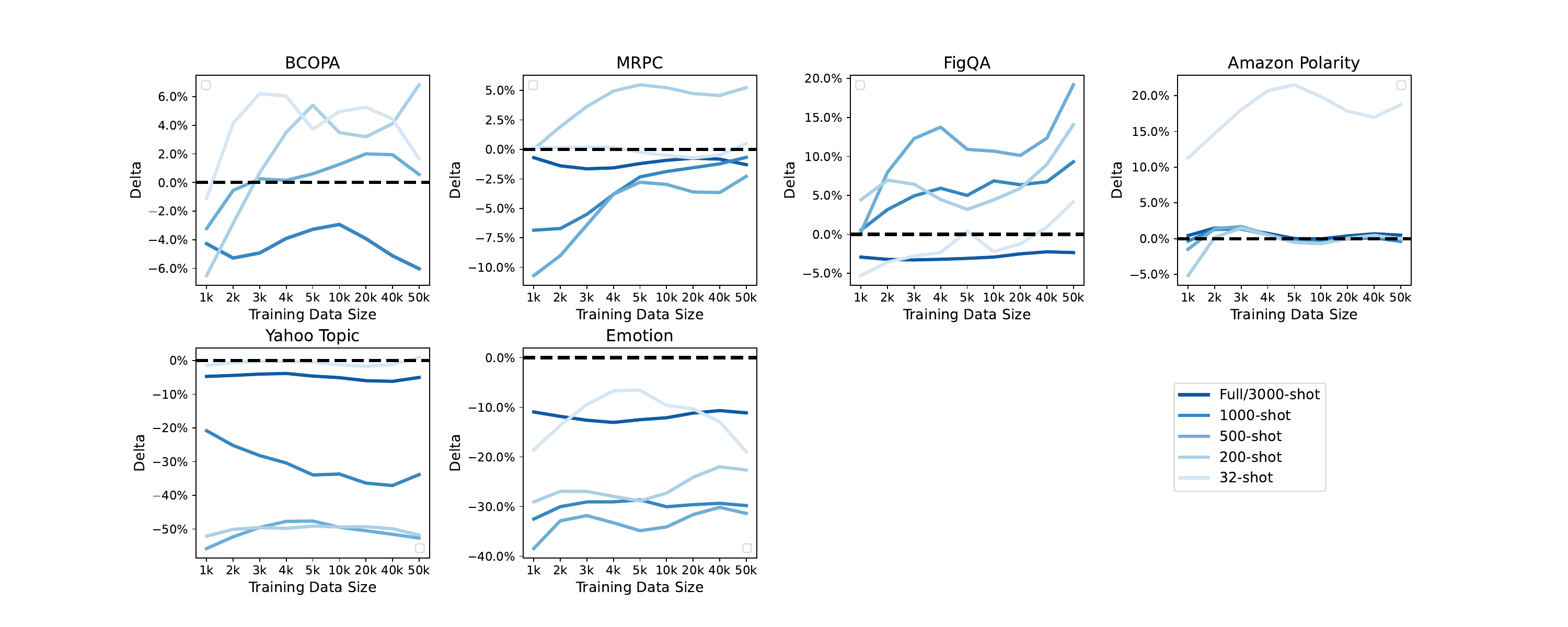}
    \caption{N-shot improvement of Statement-Tuned RoBERTa-base models used for regular finetuning. The y-axis, Delta, is the difference between the accuracy of the Statement-Tuned model fine-tuned for the task directly by discarding the \st~classification head and the accuracy achieved by regular fine-tuning of RoBERTa-base on the task. A positive Delta indicates improvement over the baseline approach.}
    \label{fig:regular-finetuning}
\end{figure*}

In figure \ref{fig:regular-finetuning}, we visualize the relative improvement of our Statement-Tuned RoBERTa-base models regularly fine-tuned on N-shot downstream data over the regularly fine-tuned RoBERTa-base. The results are not as good as fine-tuning using statements. 

\section{Task Categories Breakdown}
\label{sec:appendixe}
For the statement tuning task diversity, we group datasets based on task categories as follows (evaluation datasets are underlined):
\begin{enumerate}
    \itemsep0em
    \item Summarization (\textbf{SU}): WikiLingua, SAMSum
    \item Sentiment Analysis (\textbf{SA}): Yelp Polarity, \underline{Amazon Polarity}
    \item Question Answering (\textbf{QA}): Mintaka, SQuAD, QASC, SciQ, RACE
    \item Natural Language Inference (\textbf{NLI}): MNLI, SNLI
    \item Commonsense Reasoning (\textbf{CR}): Winogrande, PiQA
   \item Paraphrase Detection (\textbf{PD}): QQP, \underline{MRPC}
   \item Word Sense Disambiguation (\textbf{WSD}): Definite Pronoun Resolution
   \item Intent Classification (\textbf{IC}): Massive
   \item Offensive Language Identification (\textbf{OLI}): Tweet Eval's Offensive
   \item Sentence Completion: \underline{BCOPA}, \underline{StoryCloze}
   \item Emotion Recognition: \underline{Emotion}
   \item Topic Classification: \underline{Yahoo Answer Topic}
   \item Nonliteral Reasoning: \underline{FigQA}
\end{enumerate}

\section{N-shot Correlation}
\begin{figure}[!ht]
    \centering
    \includegraphics[width=\linewidth]{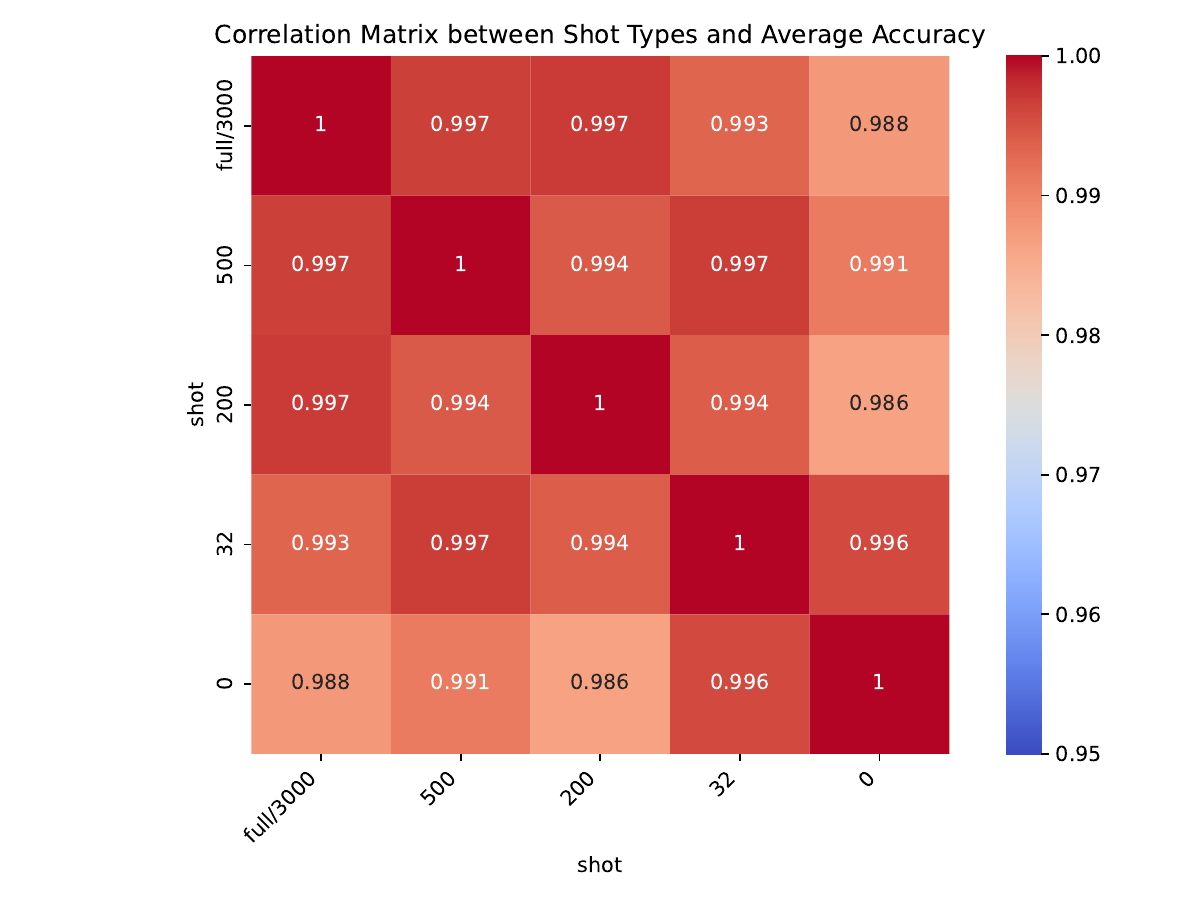}
    \caption{N-shot correlation using the average accuracy across all training set sizes and evaluation sets.}
    \label{fig:corr}
\end{figure}

Figure \ref{fig:corr} shows the correlation of accuracies achieved in different N-shot settings with various shot numbers.

\section{Inference Speed Comparison}
\label{sec:performance}
We report the average examples/sec processed for each of the datasets in Table~\ref{tab:speed}. It is important to note that all models are run on a single GPU, except for Meta-Llama-3-70B-Instruct and Llama-2-13B-chat which were run on 4 and 2 GPUs, respectively.
\begin{table*}[]
    \centering
    \resizebox{\textwidth}{!}{
    \begin{tabular}{lrrrrrrrr}
\toprule
Model & BCOPA & MRPC & FigQA & Amazon Polarity & StoryCloze & YA Topic & Emotion & Avg \\
\midrule
Qwen1.5-0.5B-Chat & 0.1 & 0.1 & 0.1 & 0.1 & 0.1 & 0.1 & 0.1 & 0.1 \\
phi-2 & 1.4 & 1.4 & 1.4 & 1.4 & 1.5 & 1.4 & 1.4 & 1.4 \\
Meta-Llama-3-70B-Instruct* & 2.9 & 1.3 & 3.2 & 0.9 & 4.9 & 1.1 & 2.1 & 2.3 \\
flan-t5-large & 8.2 & 13.2 & 13.2 & 13.2 & 13.2 & 13.2 & 13.2 & 12.5 \\
Llama-2-13b-chat-hf* & 8.7 & 5.7 & 12.8 & 4.3 & 15.7 & 4.4 & 6.9 & 8.3 \\
Our Approach (roberta-large) & 9.3 & 14.5 & 15.0 & 15.0 & 14.7 & 3.1 & 5.1 & 11.0 \\
bart-large-mnli & 9.7 & 14.1 & 14.0 & 14.2 & 14.1 & 13.7 & 13.8 & 13.4 \\
pythia-6.9b & 12.0 & 0.6 & 4.6 & 0.4 & 0.6 & 2.2 & 0.4 & 3.0 \\
Llama-2-7b-chat-hf & 12.5 & 0.6 & 4.6 & 0.4 & 0.6 & 2.3 & 0.5 & 3.1 \\
Mistral-7B-Instruct-v0.2 & 12.8 & 0.5 & 2.7 & 0.3 & 0.5 & 1.7 & 0.4 & 2.7 \\
pythia-2.8b & 13.6 & 16.7 & 24.9 & 15.2 & 27.2 & 15.1 & 20.9 & 19.1 \\
flan-t5-small & 13.9 & 39.2 & 39.1 & 39.3 & 39.4 & 39.3 & 39.3 & 35.6 \\
Our Approach (roberta-base) & 17.9 & 49.8 & 50.0 & 49.8 & 49.9 & 10.3 & 17.0 & 34.9 \\
\bottomrule
\end{tabular}}
    \caption{The average examples per second processed by each model on each task. * indicates that the model required the use of more than one GPU.}
    \label{tab:speed}
\end{table*}


\begin{table*}[]
    \centering
    \resizebox{\textwidth}{!}{
    \begin{tabular}{cccccccc}\toprule
        \textbf{} &\textbf{Llama-2-13b} &\textbf{Qwen1.5-7B} &\textbf{Pythia-6.9B} &\textbf{Pythia-2.9B} &\textbf{Phi-2} &\textbf{Qwen1.5-0.5B} \\\midrule
        \#Parameters &13B &7B &6.9B &2.9B &2.7B &500M \\
        Quantization &8bit &4bit &4bit &4bit &4bit &4bit \\
        Sequence Length &4096 &2048 &2048 &2048 &2048 &2048 \\
        lora\_r &32 &32 &32 &32 &32 &32 \\
        lora\_alpha &16 &64 &64 &64 &64 &64 \\
        lora\_dropout &0.05 &0.05 &0.05 &0.05 &0.05 &0.05 \\
        adam\_beta2 &0.999 &0.999 &0.95 &0.95 &0.95 &0.999 \\
        adam\_epsilon &1e-8 &1e-8 &0.00001 &0.00001 &0.00001 &1e-8 \\
        max\_grad\_norm &none &none &1.0 &1.0 &1.0 &none \\
        optimizer &adamw\_bnb\_8bit &adamw\_torch &adamw\_torch &adamw\_torch &adamw\_torch &adamw\_torch \\
        gradient acc. &4 &4 &4 &4 &4 &4 \\
        micro batch size &2 &1 &1 &1 &1 &1 \\
        lr\_scheduler &cosine &cosine &cosine &cosine &cosine &cosine \\
        learning\_rate &0.0002 &0.0002 &0.00001 &0.00001 &0.000003 &0.0002 \\
        \bottomrule
    \end{tabular}
    }
    \caption{Instruction Tuning Hyperparameters}
    \label{tab:ithyperparams}
\end{table*}

\section{Statement-Tuning Data Instruction-Tuned Decoders}
\label{sec:appendixi}

As seen in Table~\ref{tab:decoders-it-0shot}, we report the comparison of our approach with several decoders that were instruct-tuned using an instruction-tuning dataset created using the same training corpora used for~\st. Training details are outlined in Table~\ref{tab:ithyperparams}. Furthermore, the templates used to form instructions are based on those used for Flan \cite{DBLP:conf/iclr/WeiBZGYLDDL22} (we make the dataset available here: Anonymized Repository). 

The models tested are a subset of the ones reported in Table~\ref{tab:decoders-0shot} due to the time and computational expense of instruction-tuning and hardware limitations. However, even when the LLM models are instruction-tuned on the same data as the Statement-Tuned RoBERTa models, we observe similar trends where performance of the RoBERTa-base model tends to match the performance of all models up to 6.9B parameters on all tasks except for BCOPA. Furthermore, the RoBERTa-large model approaches or exceeds performance on all tasks for models with 7B+ parameters. The same trend of dominating performance on FigQA and StoryCloze is observed.

\begin{table*}[!ht]
    \centering
    \resizebox{\textwidth}{!}{
    \begin{tabular}[width=\linewidth]{@{ }l@{ }cccccccc@{ }|c}
    \toprule
         &  \textbf{\#Parameters} &   \multicolumn{1}{c}{\textbf{BCOPA}}   &   \multicolumn{1}{c}{\textbf{MRPC}}   &   \multicolumn{1}{c}{\textbf{FigQA}}   &   \multicolumn{1}{c}{\textbf{Amazon Polarity}}   &   \multicolumn{1}{c}{\textbf{StoryCloze}}   &   \multicolumn{1}{c}{\textbf{YA Topic}}   &   \multicolumn{1}{c}{\textbf{Emotion}} &  \multicolumn{1}{c}{\textbf{Avg}} \\
         \midrule
         Llama-2-13b &13B &89.6 &\textcolor{lightgray}{\textbf{60.8}} &\textcolor{lightgray}{\textbf{40.9}} &93.7 &82.4 &53.2 &51.6 &67.5 \\
        Qwen1.5-7B &7B &87.2 &78.9 &\textcolor{lightgray}{\textbf{41.4}} &94.8 &\textcolor{lightgray}{\textbf{75.7}} &47.8 &56.5 &68.9 \\
        Pythia-6.9B &6.9B &82.8 &\textcolor{lightgray}{\textbf{68.1}} &\textcolor{lightgray}{\textbf{40.0}} &\textcolor{lightgray}{\textbf{71.7}} &\textcolor{lightgray}{\textbf{71.5}} &\textcolor{lightgray}{\textbf{16.4}} &\textcolor{lightgray}{\textbf{27.5}} &\textcolor{lightgray}{\textbf{54.0}} \\
        Pythia-2.9B &2.9B &79.6 &\textcolor{lightgray}{\textbf{67.9}} &\textcolor{lightgray}{\textbf{40.3}} &\textcolor{lightgray}{\textbf{77.2}} &\textcolor{lightgray}{\textbf{69.7}} &\textcolor{lightgray}{\textbf{21.2}} &\textcolor{lightgray}{\textbf{30.6}} &\textcolor{lightgray}{\textbf{55.2}} \\
        Phi-2 &2.7B &87.2 &\textcolor{lightgray}{\textbf{68.1}} &\textcolor{lightgray}{\textbf{41.7}} &\textcolor{lightgray}{\textbf{85.6}} &\textcolor{lightgray}{\textbf{77.8}} &\textcolor{lightgray}{\textbf{38.4}} &53.5 &\textcolor{lightgray}{\textbf{64.6}} \\
        Qwen1.5-0.5B &500M &\textcolor{lightgray}{\textbf{72.4}} &\textcolor{lightgray}{\textbf{68.4}} &\textcolor{lightgray}{\textbf{39.4}} &\textcolor{lightgray}{\textbf{49.8}} &\textcolor{lightgray}{\textbf{67.6}} &\textcolor{lightgray}{\textbf{33.2}} &72.4 &\textcolor{lightgray}{\textbf{57.6}} \\
    \hdashline
    \textbf{Our Approach: RoBERTa-base (Best)} &\textbf{125M} &75.3$_{(0.5)}$ &72.3$_{(1.5)}$ &61.4$_{(0.6)}$ &92.9$_{(1.3)}$ &79.1$_{(1.1)}$ &40.2$_{(3.8)}$ &48.5$_{(5.1)}$ & 67.1 \\
    \textbf{Our Approach: RoBERTa-base (4k)} &\textbf{125M} &72.4$_{(0.5)}$ &69.6$_{(1.1)}$ &60.7$_{(0.9)}$ &92.3$_{(0.8)}$ &78.5$_{(2.7)}$ & 37.9$_{(2.7)}$ &46.6$_{(4.3)}$ & 65.4 \\
    \textbf{Our Approach: RoBERTa-large (Best)} &\textbf{355M} &85.1$_{(0.7)}$ &71.8$_{(0.8)}$ &74.2$_{(1.4)}$ &95.4$_{(0.4)}$ & 92.1$_{(0.7)}$ &49.9$_{(2.1)}$ &50.7$_{(1.4)}$ & 75.3 \\
    \textbf{Our Approach: RoBERTa-large (10k)} &\textbf{355M} &85.1$_{(0.7)}$ &71.5$_{(0.8)}$ &73.0$_{(2.4)}$ &95.4$_{(0.4)}$ &91.1$_{(0.8)}$ &48.4$_{(0.7)}$ &49.1$_{(3.2)}$ & 73.4 \\
    \midrule
    \multicolumn{6}{l}{\textbf{Full/3000-shot:}}\\
    RoBERTa-base (FT) &125M &\multicolumn{1}{c}{74.2} &\multicolumn{1}{c}{87.0} &\multicolumn{1}{c}{88.1} &\multicolumn{1}{c}{94.3} &\multicolumn{1}{c}{-} &\multicolumn{1}{c}{71.0} &\multicolumn{1}{c}{82.2} & \multicolumn{1}{c}{-} \\
    RoBERTa-large (FT) &355M &\multicolumn{1}{c}{86.0} &\multicolumn{1}{c}{87.6} &\multicolumn{1}{c}{92.0} &\multicolumn{1}{c}{96.5} &\multicolumn{1}{c}{-} &\multicolumn{1}{c}{68.5} &\multicolumn{1}{c}{78.2}& \multicolumn{1}{c}{-} \\
    \bottomrule
    \end{tabular}
}
    \caption{Comparison of our approach against many pretrained open-source encoder-Decoder and Decoder-only Instruction-tuned Pretrained Large Language Models on 7 Natural Language Understanding tasks in Zero-shot conditions. FT stands for Full Finetuning and is included for reference. For \st, we report the average across 5 training runs and 5 evaluation runs and include the average standard deviation in parenthesis. We highlight all scores in gray where our approach with RoBERTa-base (best) exceeds or is equal to the score given by the model.}
    \label{tab:decoders-it-0shot}
\end{table*}

\end{document}